\UseRawInputEncoding%For ARXIV
\documentclass[runningheads]{llncs}

\usepackage{booktabs}
\usepackage{multirow}
\usepackage{multicol}
\pdfoutput=1%For ARXIV

\usepackage[T1]{fontenc}
\usepackage{graphicx}
\begin{document}
\title{Optimizing Deep Transformers for Chinese-Thai Low-Resource Translation}
%
%\titlerunning{Abbreviated paper title}
% If the paper title is too long for the running head, you can set
% an abbreviated paper title here
%

\author{Wenjie Hao\inst{1} \orcidID{0000-0002-8961-7052} \and
Hongfei Xu\inst{1} \and
Lingling Mu\inst{1} \and
Hongying Zan\inst{1,2}
}
\authorrunning{Hao et al.}
% First names are abbreviated in the running head.
% If there are more than two authors, 'et al.' is used.
%
\institute{
Zhengzhou University, Henan 450001, China\\
\email{haowj9977@163.com,hfxunlp@foxmail.com,\{iellmu,iehyzan\}@zzu.edu.cn}\\
\and
Peng Cheng Laboratory, Shenzhen 518000, China
}

\if false
\author{First Author\inst{1}\orcidID{0000-1111-2222-3333} \and
Second Author\inst{2,3}\orcidID{1111-2222-3333-4444} \and
Third Author\inst{3}\orcidID{2222--3333-4444-5555}}
\authorrunning{F. Author et al.}
% First names are abbreviated in the running head.
% If there are more than two authors, 'et al.' is used.
%
\institute{Princeton University, Princeton NJ 08544, USA \and
Springer Heidelberg, Tiergartenstr. 17, 69121 Heidelberg, Germany
\email{lncs@springer.com}\\
\url{http://www.springer.com/gp/computer-science/lncs} \and
ABC Institute, Rupert-Karls-University Heidelberg, Heidelberg, Germany\\
\email{\{abc,lncs\}@uni-heidelberg.de}}
\fi

\maketitle              % typeset the header of the contribution
\begin{abstract}
In this paper, we study the use of deep Transformer translation model for the CCMT 2022 Chinese$\leftrightarrow$Thai low-resource machine translation task. We first explore the experiment settings (including the number of BPE merge operations, dropout probability, embedding size, etc.) for the low-resource scenario with the 6-layer Transformer. Considering that increasing the number of layers also increases the regularization on new model parameters (dropout modules are also introduced when using more layers), we adopt the highest performance setting but increase the depth of the Transformer to 24 layers to obtain improved translation quality. Our work obtains the SOTA performance in the Chinese-to-Thai translation in the constrained evaluation.

\end{abstract}

\section{Introduction}

Neural machine translation (NMT) has achieved impressive performance with the support of large amounts of parallel data \cite{NIPS2017_attention,akhbardeh-EtAl:2021:WMT}. However, in low-resource scenario, its performance is far from expectation \cite{koehn-knowles-2017-six,lample-etal-2018-phrase}.

To improve the translation performance, previous work either study data augmentation approaches to leverage pseudo data \cite{sennrich-etal-2016-improving,edunov-etal-2018-understanding,fadaee-etal-2017-data,wang-etal-2018-switchout,mallinson-etal-2017-paraphrasing} or benefit from models pre-trained on large-scale monolingual corpus \cite{neishi-etal-2017-bag,qi-etal-2018-pre}.

Instead of introducing more data, in this paper, we explore the effects of different data processing and model settings for the CCMT 2022 Chinese$\leftrightarrow$Thai low-resource machine translation task inspired by Sennrich and Zhang \cite{sennrich-zhang-2019-revisiting}.

Specifically, we adopt the  Chinese$\leftrightarrow$Thai (Zh$\leftrightarrow$Th) machine translation data from CCMT 2022 of 200k training sentence pairs. We first apply strict rules for data cleaning, and employ the cutting-edge Transformer model \cite{NIPS2017_attention}. We explore the influence of BPE merge operations on performance, and the effects of different model settings (embedding size, dropout probability). As previous work \cite{bapna2018training,wang2019learning,wu2019depth,wei2020multiscale,zhang2019improving,xu-etal-2020-lipschitz,li2020shallow,huang2020improving,xiong2020on,mehta2021delight,li2021learning,xu-etal-2021-optimizing,DBLP:journals/corr/abs-1909-11942} shows that deep Transformers can bring about improved translation performance, we adopt the setting of highest performance in ablation experiments for the Chinese$\leftrightarrow$Thai translation task but increase the number of layers to 24. We explore experiment settings with the 6-layer setting but adopt the best one to deeper models, because: 1) exploring the effects of these hyper-parameters with shallow models is more computation-friendly than with deep models, and 2) increasing the number of layers also introduces regularization, as adding new layers also brings dropout modules.

\section{Background}
\subsection{Transformer}
Vaswani et al. \cite{NIPS2017_attention} propose the self-attention based Transformer model, evading the parallelization issue of RNN. Transformer has become the most popular model in NMT field. Transformer consists of one encoder and one decoder module, each of them is  formed by several layers, and the multi-layer structure allows it to model complicated functions. The Transformer model also employs residual connection and layer normalization techniques for the purpose of ease optimization.

\subsection{Low-Resource NMT}
Despite that NMT has achieved impressive performance in high-resource cases \cite{NIPS2017_attention}, its performance drops heavily in low-resource scenarios, even under-performing phrase-based statistical machine translation (PBSMT) \cite{koehn-knowles-2017-six,lample-etal-2018-phrase}. NMT normally requires large amounts of auxiliary data to achieve competitive results. Sennrich and Zhang \cite{sennrich-zhang-2019-revisiting} show that this is due to the lack of system adaptation for low-resource settings. They suggest that large vocabularies lead to low-frequency (sub)words, and the amount of data is not sufficient to learn high-quality high-dimensional representations for these low-frequency tokens. Reducing the vocabulary size (14k$\rightarrow$2k symbols) can bring significant improvements (7.20$\rightarrow$12.10 BLEU). In addition, they show that aggressive (word) dropout (0.1$\rightarrow$0.3) can bring impressive performance (13.03$\rightarrow$15.84 BLEU), and reducing batch size (4k$\rightarrow$1k tokens) may also benefit. Optimized NMT systems can indeed outperform PBSMT. 

\subsection{Parameter Initialization for Deep Transformers}
Xu et al. \cite{xu-etal-2020-lipschitz} suggest that the training issue of deep Transformers is because that the layer normalization may shrink the residual connections, leading to the gradient vanishing issue. They propose to address this by applying the Lipschitz constraint to parameter initialization. Experiments on WMT14 English-German and WMT15 Czech-English translation tasks show the effectiveness of their simple approach.

\subsection{Deep Transformers for Low-Resource Tasks}
Previous work shows that deep Transformers generally perform well with sufficient training data \cite{DBLP:journals/corr/abs-1909-11942},  and few attempts have been made on training deep Transformers from scratch on small datasets. Xu et al. \cite{xu-etal-2021-optimizing} propose Data-dependent Transformer Fixed-update initialization scheme, called DT-Fixup,  and experiment on the Text-to-SQL semantic parsing and the logical reading comprehension tasks. They show that deep Transformers can work better than their shallow counterparts on small datasets through proper initialization and optimization procedure. Their work inspires us to explore the use of deep Transformers for low-resource machine translation.

\section{Our Work}
\label{sec:approach}

\subsection{Data Processing}
\label{process:3.2}

The quality of the dataset affects the performance of NMT. Therefor, We first standardize the texts with the following pipeline:

\begin{enumerate}
 \item removing sentences with encoding errors;
 \item converting Traditional Chinese to Simplified Chinese through OpenCC; \footnote{\url{https://github.com/BYVoid/OpenCC}}
 \item replacing full width characters with their corresponding half width characters;
 \item converting all named and numeric character HTML references (e.g., \&gt;, \&\#62;, \&\#x3e) to the corresponding Unicode characters.
\end{enumerate}

For the training of NMT models, we  segment Chinese sentences into words using jieba. \footnote{\url{https://github.com/fxsjy/jieba}}

We perform independent Byte Pair Encoding (BPE) \cite{sennrich-etal-2016-neural} for Thai and Chinese corpus  to address the unknown word issue with the SentencePiece toolkit \cite{kudo-richardson-2018-sentencepiece}.
\footnote{\url{https://github.com/google/sentencepiece}}

As the evaluation does not
release the test set, we hold out the last 1000 sentence pairs of the training set for validation.

\subsection{Exploration of Training Settings}
We explore the influence of different training settings on the low-resource translation task in two aspects:
\begin{enumerate}
\item Vocabulary sizes;  
\item Model hyper-parameters (embedding size and dropout probabilities). % .  
\end{enumerate}

For our experiment, we employ the Transformer translation model \cite{NIPS2017_attention} for NMT, as it has achieved the state-of-the-art performance in MT evaluations \cite{akhbardeh-EtAl:2021:WMT} and conduct our experiment based on the Neutron toolkit \cite{Neutron} system. Neutron is an  open source Transformer \cite{NIPS2017_attention} implementation of the Transformer and its variants based on PyTorch.

\subsubsection{Exploration of vocabulary sizes}
 
Previous work shows that the effect of vocabulary size on translation quality is relatively small for high-resource settings \cite{haddow-etal-2018-university}. While for low-resource settings, reduced vocabulary size (14k$\rightarrow$2k) may benefit translation quality  \cite{sennrich-zhang-2019-revisiting}. BPE \cite{sennrich-etal-2016-neural} is a popular choice for open-vocabulary translation, which has one hyper-parameter, the number of merge operations, that  determines the final vocabulary size. Following Sennrich and Zhang \cite{sennrich-zhang-2019-revisiting}, we explore the influence of different vocabulary sizes for the Thai$\rightarrow$Chinese translation task.

We train 4 NMT models in Thai$\rightarrow$Chinese translation direction with different number of BPE merge operations, and the statistics of resulted vocabularies are shown in Table~\ref{tab:vocab-settings}. Specifically, we perform independent BPE \cite{sennrich-etal-2016-neural} for Thai and Chinese corpus with 4k/8k/16k/24k merge operations by SentencePiece \cite{kudo-richardson-2018-sentencepiece}. 
%\footnote{\url{https://github.com/google/sentencepiece}}.

For model settings, we adopted the Transformer with 6 encoder and decoder layers, 256 as the embedding dimension and 4 times of embedding dimension as the number of hidden units of the feed-forward layer, a dropout  probability of 0.1. We used relative position \cite{shaw-etal-2018-self} with a clipping distance k of 16. The number of warm-up steps was set to $8k$. We used a batch size of around $25k$ target tokens achieved by gradient accumulation, and trained the models for 128 epochs.

For evaluation, we decode with a beam size of 4 with average of the last $5$ checkpoints saved in an interval of $1,500$
training steps. We evaluate the translation quality by character BLEU with the SacreBLEU toolkit \cite{post-2018-call}.  Results are shown in Table~\ref{tab:vocab-settings}.

Table~\ref{tab:vocab-settings} shows that:  1) in general, the use of more merge operations (16k/24k) is better than fewer ones (6k/8k), and 2) the setting of 16k merge operations leads to the best performance for the Thai$\rightarrow$Chinese translation task, achieving 29.90 BLEU points.  

% vocab
\begin{table}[t]
\small
 \centering
 \caption{Results (BLEU) on CCMT 2022 Th$\rightarrow$Zh translation task with different vocabulary sizes.}
 \begin{tabular}{l|cccc}
 \toprule
Merge operations & 6k &8k & 16k & 24k  \\
 \midrule
 Thai vocabulary size &5,996 & 7,997 & 16,000 &23,999\\
 Chinese vocabulary size &5,989 & 7,984 & 15,943 &23,881 \\
 \midrule
 BLEU  &27.07 & 25.70 & \textbf{29.90} & 28.87 \\
 \bottomrule
 \end{tabular}%
 
 \label{tab:vocab-settings}%
\end{table}%

\subsubsection{Exploration of  hyper-parameter settings of model}
Hyper-parameters are often re-used across experiments. However, best practices may differ between high-resource and low-resource settings. While the trend in high-resource settings is using large and deep models, Nguyen and Chiang \cite{nguyen-chiang-2018-improving} use small models with fewer layers for small datasets, and Sennrich and Zhang \cite{sennrich-zhang-2019-revisiting} show that aggressive dropout is better for low-resource translation. In this paper, we also explore the effects of model sizes (embedding dimension and hidden dimension) and dropout probabilities on the performance. 

We train 5 NMT models in Chinese$\rightarrow$Thai translation direction with different training settings as shown in Table~\ref{tab:settings}. We set the number of BPE merge operations to 16k based on Table~\ref{tab:vocab-settings}.

We experimented the Transformers with 6 encoder and decoder layers, 256 / 384 / 512 as the embedding dimension and 4 times of embedding dimension as the number of hidden units of the feed-forward layer, dropout  probabilities of 0.1 or 0.3.  We used relative position \cite{shaw-etal-2018-self} with a clipping distance k was 16 and GeLU as the activation function. The number of warm-up steps was set to $8k$. We used a batch size of around $25k$ target tokens achieved by gradient accumulation, and trained the models for 128 epochs.

For evaluation, we decode with a beam size of 4, and evaluate the translation quality with the SacreBLEU toolkit \cite{post-2018-call} with the average of the last $5$ checkpoints saved in an interval of $1,500$ training steps. Results are shown in Table~\ref{tab:settings}.

Table~\ref{tab:settings} shows that: 1)  large embedding dimension is beneficial to translation performance, 2) aggressive dropout (0.3 in this paper) does not benefit the task, and 3) Setting D with 512 as the embedding dimension and 0.1 as the dropout probability is the best option, achieving a  BLEU score of 24.42 in the Chinese$\rightarrow$Thai translation task.

% Table generated by Excel2LaTeX from sheet 'pingce'
\begin{table}[t]
  \centering
  \caption{Results (BLEU) on CCMT 2022 Zh$\rightarrow$Th translation task with different model settings.}
    \begin{tabular}{lcccccc}
    \toprule
    \multicolumn{2}{l}{Settings} & \multicolumn{1}{l}{A} & \multicolumn{1}{l}{B} & \multicolumn{1}{l}{C} & \multicolumn{1}{l}{D} & \multicolumn{1}{l}{E} \\
    \midrule
    \multirow{3}[0]{*}{Embbeding size} & \multicolumn{1}{p{4em}}{256} & \multicolumn{1}{l}{$\surd$} &       &       &       &  \\
          & \multicolumn{1}{p{4em}}{384} &       & \multicolumn{1}{l}{$\surd$} & \multicolumn{1}{l}{$\surd$} &       &  \\
          & \multicolumn{1}{p{4em}}{512} &       &       &       & \multicolumn{1}{l}{$\surd$} & \multicolumn{1}{l}{$\surd$} \\
          \midrule
    \multirow{2}[0]{*}{Dropout} & \multicolumn{1}{{p{4em}}}{0.1} & \multicolumn{1}{l}{$\surd$} & \multicolumn{1}{l}{$\surd$} &       & \multicolumn{1}{l}{$\surd$} &  \\
          & \multicolumn{1}{{p{4em}}}{0.3} &       &       & \multicolumn{1}{l}{$\surd$} &       & \multicolumn{1}{l}{$\surd$} \\
          \midrule
    \multicolumn{2}{l}{BLEU} & 6.35  & 15.02 & 5.30   & \textbf{24.42} & 7.73 \\
    \bottomrule
    \end{tabular}%
  \label{tab:settings}%
\end{table}%

\subsection{Deep Transformers for Low-Resource Machine Translation }
To obtain good translation quality,  we adopt the setting D, but use 24 encoder and decoder layers for better performance \cite{bapna2018training,wang2019learning,wu2019depth,wei2020multiscale,zhang2019improving,xu-etal-2020-lipschitz,li2020shallow,huang2020improving,xiong2020on,mehta2021delight,li2021learning,xu-etal-2021-optimizing,DBLP:journals/corr/abs-1909-11942}. Parameters were initialized under the Lipschitz constraint \cite{xu-etal-2020-lipschitz} to ensure the convergence. We used the dynamical batch size strategy which dynamically determines proper and efficient batch sizes during training \cite{xu-etal-2020-dynamically}. 

We use the best experiment setting explored with the 6-layer models for deeper models, because: 1) training shallow models are much faster than deep models, and 2) adding new layers also introduces regularization, as dropout modules are also introduced with these layers.

We train two models on the whole training set, which takes about 75 hours to train one model on a nvidia RTX3090 GPU. We averaged the last $20$ checkpoints saved with an interval of $1,500$ training steps.

We decode the CCMT 2022 Zh$\leftrightarrow$Th test set consisting of 10k sentences for each direction with a beam size of 4. Results are shown in Table~\ref{tab:last}.

Table~\ref{tab:last} shows that the CCMT 2022 Chinese-Thai low-resource translation task is still a quite challenging task and there is a quite large space for improvements. But to date, our study establishes the SOTA performance in the Chinese-to-Thai translation in the constrained evaluation.

% Table generated by Excel2LaTeX from sheet 'pingce'
\begin{table}[t]
  \centering
  \caption{Results on the CCMT 2022 Zh$\leftrightarrow$Th test set. The computation of BLEU scores for the test set are different from that for Tables \ref{tab:vocab-settings} and \ref{tab:settings}.}
    \begin{tabular}{lcc}
          \toprule
          & \multicolumn{1}{l}{Th$\rightarrow$Zh} & \multicolumn{1}{l}{Zh$\rightarrow$Th} \\
          \midrule
    BLEU4 &  /     & 9.06 \\
    BLEU5 & 4.85  & / \\
    \bottomrule
    \end{tabular}%
  \label{tab:last}%
\end{table}%

\section{Related Work}
 
As data scarcity is the main problem of low-resource machine translation,  making most of the existing data is a popular research direction to address this issue in previous work. There are two specific types: 1) data augmentation, and 2) using pre-trained language models. 

Data augmentation is to add training data, normally through modifications of existing data or the generation of new pseudo data. In machine translation, typical data enhancement methods include back-translating external monolingual data \cite{sennrich-etal-2016-improving,edunov-etal-2018-understanding}, obtaining pseudo bilingual data by modifying original bilingual data, such as adding noise to training data \cite{fadaee-etal-2017-data,wang-etal-2018-switchout} or by paraphrasing which takes into the diversity of natural language expression into account
\cite{mallinson-etal-2017-paraphrasing}, and mining of bilingual sentence pairs from comparable corpus  \cite{wu-etal-2019-machine} (comparable corpus is a text that is not fully translated from the source language to the target language but contains with rich knowledge of bilingual contrast).

For the use of pre-trained language models in NMT, leveraging the target-side language model is the most straightforward way to use monolingual data \cite{stahlberg-etal-2018-simple}. Other work \cite{neishi-etal-2017-bag,qi-etal-2018-pre} directly uses word embeddings  pre-trained on monolingual data to initialize the word embedding matrix of NMT models. More recently, some studies leverage pre-trained models to initialize the model parameters of the encoder of NMT \cite{clinchant-etal-2019-use,imamura-sumita-2019-recycling,edunov-etal-2019-pre}.

Fore-mentioned studies require large amounts of auxiliary data. Low-resource NMT without auxiliary data has received comparably less attention \cite{nguyen-chiang-2018-improving,sennrich-zhang-2019-revisiting}. In this work, we revisit this point with deep Transformers, and focus on techniques to adapt deep Transformers to make most of low-resource parallel training data, exploring the vocabulary sizes and model settings for NMT.

\section{Conclusion}
In this paper, we explore the influence of different settings for the use of deep Transformers on the CCMT 2022 Zh$\leftrightarrow$Th low-resource translation task.

We first test the effects of the number of BPE merge operations, embedding dimension and dropout probabilities with 6-layer models, then adapt the best setting to the 24-layer model, under the motivation that: 1) shallow models are fast to train, and 2) increasing the number of layers also introduces regularization for these added layers.

\section*{Acknowledgments}
We thank anonymous reviewers for their insightful comments. We acknowledge the support of the National Social Science Fund of China (Grant No. 17ZDA138 and Grant No. 14BYY096).

%
% ---- Bibliography ----
%
% BibTeX users should specify bibliography style 'splncs04'.
% References will then be sorted and formatted in the correct style.
%
\bibliographystyle{splncs04}
\bibliography{anthology,custom}

\begin{thebibliography}{10}
\providecommand{\url}[1]{\texttt{#1}}
\providecommand{\urlprefix}{URL }
\providecommand{\doi}[1]{https://doi.org/#1}

\bibitem{akhbardeh-EtAl:2021:WMT}
Akhbardeh, F., Arkhangorodsky, A., Biesialska, M., Bojar, O., Chatterjee, R.,
  Chaudhary, V., Costa-jussa, M.R., España-Bonet, C., Fan, A., Federmann, C.,
  Freitag, M., Graham, Y., Grundkiewicz, R., Haddow, B., Harter, L., Heafield,
  K., Homan, C., Huck, M., Amponsah-Kaakyire, K., Kasai, J., Khashabi, D.,
  Knight, K., Kocmi, T., Koehn, P., Lourie, N., Monz, C., Morishita, M.,
  Nagata, M., Nagesh, A., Nakazawa, T., Negri, M., Pal, S., Tapo, A.A., Turchi,
  M., Vydrin, V., Zampieri, M.: Findings of the 2021 conference on machine
  translation (wmt21). In: Proceedings of the Sixth Conference on Machine
  Translation. pp. 1--88. Association for Computational Linguistics, Online
  (November 2021), \url{https://aclanthology.org/2021.wmt-1.1}

\bibitem{bapna2018training}
Bapna, A., Chen, M., Firat, O., Cao, Y., Wu, Y.: Training deeper neural machine
  translation models with transparent attention. In: Proceedings of the 2018
  Conference on Empirical Methods in Natural Language Processing. pp.
  3028--3033. Association for Computational Linguistics (2018),
  \url{http://aclweb.org/anthology/D18-1338}

\bibitem{clinchant-etal-2019-use}
Clinchant, S., Jung, K.W., Nikoulina, V.: On the use of {BERT} for neural
  machine translation. In: Proceedings of the 3rd Workshop on Neural Generation
  and Translation. pp. 108--117. Association for Computational Linguistics,
  Hong Kong (Nov 2019). \doi{10.18653/v1/D19-5611},
  \url{https://aclanthology.org/D19-5611}

\bibitem{edunov-etal-2019-pre}
Edunov, S., Baevski, A., Auli, M.: Pre-trained language model representations
  for language generation. In: Proceedings of the 2019 Conference of the North
  {A}merican Chapter of the Association for Computational Linguistics: Human
  Language Technologies, Volume 1 (Long and Short Papers). pp. 4052--4059.
  Association for Computational Linguistics, Minneapolis, Minnesota (Jun 2019).
  \doi{10.18653/v1/N19-1409}, \url{https://aclanthology.org/N19-1409}

\bibitem{edunov-etal-2018-understanding}
Edunov, S., Ott, M., Auli, M., Grangier, D.: Understanding back-translation at
  scale. In: Proceedings of the 2018 Conference on Empirical Methods in Natural
  Language Processing. pp. 489--500. Association for Computational Linguistics,
  Brussels, Belgium (Oct-Nov 2018). \doi{10.18653/v1/D18-1045},
  \url{https://aclanthology.org/D18-1045}

\bibitem{fadaee-etal-2017-data}
Fadaee, M., Bisazza, A., Monz, C.: Data augmentation for low-resource neural
  machine translation. In: Proceedings of the 55th Annual Meeting of the
  Association for Computational Linguistics (Volume 2: Short Papers). pp.
  567--573. Association for Computational Linguistics, Vancouver, Canada (Jul
  2017). \doi{10.18653/v1/P17-2090}, \url{https://aclanthology.org/P17-2090}

\bibitem{haddow-etal-2018-university}
Haddow, B., Bogoychev, N., Emelin, D., Germann, U., Grundkiewicz, R., Heafield,
  K., Miceli~Barone, A.V., Sennrich, R.: The {U}niversity of {E}dinburgh{'}s
  submissions to the {WMT}18 news translation task. In: Proceedings of the
  Third Conference on Machine Translation: Shared Task Papers. pp. 399--409.
  Association for Computational Linguistics, Belgium, Brussels (Oct 2018).
  \doi{10.18653/v1/W18-6412}, \url{https://aclanthology.org/W18-6412}

\bibitem{huang2020improving}
Huang, X.S., Perez, F., Ba, J., Volkovs, M.: Improving transformer optimization
  through better initialization. In: III, H.D., Singh, A. (eds.) Proceedings of
  the 37th International Conference on Machine Learning. Proceedings of Machine
  Learning Research, vol.~119, pp. 4475--4483. PMLR (13--18 Jul 2020),
  \url{https://proceedings.mlr.press/v119/huang20f.html}

\bibitem{imamura-sumita-2019-recycling}
Imamura, K., Sumita, E.: Recycling a pre-trained {BERT} encoder for neural
  machine translation. In: Proceedings of the 3rd Workshop on Neural Generation
  and Translation. pp. 23--31. Association for Computational Linguistics, Hong
  Kong (Nov 2019). \doi{10.18653/v1/D19-5603},
  \url{https://aclanthology.org/D19-5603}

\bibitem{koehn-knowles-2017-six}
Koehn, P., Knowles, R.: Six challenges for neural machine translation. In:
  Proceedings of the First Workshop on Neural Machine Translation. pp. 28--39.
  Association for Computational Linguistics, Vancouver (Aug 2017).
  \doi{10.18653/v1/W17-3204}, \url{https://aclanthology.org/W17-3204}

\bibitem{kudo-richardson-2018-sentencepiece}
Kudo, T., Richardson, J.: {S}entence{P}iece: A simple and language independent
  subword tokenizer and detokenizer for neural text processing. In: Proceedings
  of the 2018 Conference on Empirical Methods in Natural Language Processing:
  System Demonstrations. pp. 66--71. Association for Computational Linguistics,
  Brussels, Belgium (Nov 2018). \doi{10.18653/v1/D18-2012},
  \url{https://aclanthology.org/D18-2012}

\bibitem{lample-etal-2018-phrase}
Lample, G., Ott, M., Conneau, A., Denoyer, L., Ranzato, M.: Phrase-based {\&}
  neural unsupervised machine translation. In: Proceedings of the 2018
  Conference on Empirical Methods in Natural Language Processing. pp.
  5039--5049. Association for Computational Linguistics, Brussels, Belgium
  (Oct-Nov 2018). \doi{10.18653/v1/D18-1549},
  \url{https://aclanthology.org/D18-1549}

\bibitem{DBLP:journals/corr/abs-1909-11942}
Lan, Z., Chen, M., Goodman, S., Gimpel, K., Sharma, P., Soricut, R.: {ALBERT:}
  {A} lite {BERT} for self-supervised learning of language representations.
  CoRR  \textbf{abs/1909.11942} (2019), \url{http://arxiv.org/abs/1909.11942}

\bibitem{li2021learning}
Li, B., Wang, Z., Liu, H., Du, Q., Xiao, T., Zhang, C., Zhu, J.: Learning
  light-weight translation models from deep transformer. Proceedings of the
  AAAI Conference on Artificial Intelligence  \textbf{35}(15),  13217--13225
  (May 2021), \url{https://ojs.aaai.org/index.php/AAAI/article/view/17561}

\bibitem{li2020shallow}
Li, B., Wang, Z., Liu, H., Jiang, Y., Du, Q., Xiao, T., Wang, H., Zhu, J.:
  Shallow-to-deep training for neural machine translation. In: Proceedings of
  the 2020 Conference on Empirical Methods in Natural Language Processing
  (EMNLP). pp. 995--1005. Association for Computational Linguistics, Online
  (Nov 2020). \doi{10.18653/v1/2020.emnlp-main.72},
  \url{https://aclanthology.org/2020.emnlp-main.72}

\bibitem{mallinson-etal-2017-paraphrasing}
Mallinson, J., Sennrich, R., Lapata, M.: Paraphrasing revisited with neural
  machine translation. In: Proceedings of the 15th Conference of the {E}uropean
  Chapter of the Association for Computational Linguistics: Volume 1, Long
  Papers. pp. 881--893. Association for Computational Linguistics, Valencia,
  Spain (Apr 2017), \url{https://aclanthology.org/E17-1083}

\bibitem{mehta2021delight}
Mehta, S., Ghazvininejad, M., Iyer, S., Zettlemoyer, L., Hajishirzi, H.:
  Delight: Deep and light-weight transformer. In: International Conference on
  Learning Representations (2021),
  \url{https://openreview.net/forum?id=ujmgfuxSLrO}

\bibitem{neishi-etal-2017-bag}
Neishi, M., Sakuma, J., Tohda, S., Ishiwatari, S., Yoshinaga, N., Toyoda, M.: A
  bag of useful tricks for practical neural machine translation: Embedding
  layer initialization and large batch size. In: Proceedings of the 4th
  Workshop on {A}sian Translation ({WAT}2017). pp. 99--109. Asian Federation of
  Natural Language Processing, Taipei, Taiwan (Nov 2017),
  \url{https://aclanthology.org/W17-5708}

\bibitem{nguyen-chiang-2018-improving}
Nguyen, T., Chiang, D.: Improving lexical choice in neural machine translation.
  In: Proceedings of the 2018 Conference of the North {A}merican Chapter of the
  Association for Computational Linguistics: Human Language Technologies,
  Volume 1 (Long Papers). pp. 334--343. Association for Computational
  Linguistics, New Orleans, Louisiana (Jun 2018). \doi{10.18653/v1/N18-1031},
  \url{https://aclanthology.org/N18-1031}

\bibitem{post-2018-call}
Post, M.: A call for clarity in reporting {BLEU} scores. In: Proceedings of the
  Third Conference on Machine Translation: Research Papers. pp. 186--191.
  Association for Computational Linguistics, Brussels, Belgium (Oct 2018).
  \doi{10.18653/v1/W18-6319}, \url{https://aclanthology.org/W18-6319}

\bibitem{qi-etal-2018-pre}
Qi, Y., Sachan, D., Felix, M., Padmanabhan, S., Neubig, G.: When and why are
  pre-trained word embeddings useful for neural machine translation? In:
  Proceedings of the 2018 Conference of the North {A}merican Chapter of the
  Association for Computational Linguistics: Human Language Technologies,
  Volume 2 (Short Papers). pp. 529--535. Association for Computational
  Linguistics, New Orleans, Louisiana (Jun 2018). \doi{10.18653/v1/N18-2084},
  \url{https://aclanthology.org/N18-2084}

\bibitem{sennrich-etal-2016-improving}
Sennrich, R., Haddow, B., Birch, A.: Improving neural machine translation
  models with monolingual data. In: Proceedings of the 54th Annual Meeting of
  the Association for Computational Linguistics (Volume 1: Long Papers). pp.
  86--96. Association for Computational Linguistics, Berlin, Germany (Aug
  2016). \doi{10.18653/v1/P16-1009}, \url{https://aclanthology.org/P16-1009}

\bibitem{sennrich-etal-2016-neural}
Sennrich, R., Haddow, B., Birch, A.: Neural machine translation of rare words
  with subword units. In: Proceedings of the 54th Annual Meeting of the
  Association for Computational Linguistics (Volume 1: Long Papers). pp.
  1715--1725. Association for Computational Linguistics, Berlin, Germany (Aug
  2016). \doi{10.18653/v1/P16-1162}, \url{https://aclanthology.org/P16-1162}

\bibitem{sennrich-zhang-2019-revisiting}
Sennrich, R., Zhang, B.: Revisiting low-resource neural machine translation: A
  case study. In: Proceedings of the 57th Annual Meeting of the Association for
  Computational Linguistics. pp. 211--221. Association for Computational
  Linguistics, Florence, Italy (Jul 2019). \doi{10.18653/v1/P19-1021},
  \url{https://aclanthology.org/P19-1021}

\bibitem{shaw-etal-2018-self}
Shaw, P., Uszkoreit, J., Vaswani, A.: Self-attention with relative position
  representations. In: Proceedings of the 2018 Conference of the North
  {A}merican Chapter of the Association for Computational Linguistics: Human
  Language Technologies, Volume 2 (Short Papers). pp. 464--468. Association for
  Computational Linguistics, New Orleans, Louisiana (Jun 2018).
  \doi{10.18653/v1/N18-2074}, \url{https://aclanthology.org/N18-2074}

\bibitem{stahlberg-etal-2018-simple}
Stahlberg, F., Cross, J., Stoyanov, V.: Simple fusion: Return of the language
  model. In: Proceedings of the Third Conference on Machine Translation:
  Research Papers. pp. 204--211. Association for Computational Linguistics,
  Brussels, Belgium (Oct 2018). \doi{10.18653/v1/W18-6321},
  \url{https://aclanthology.org/W18-6321}

\bibitem{NIPS2017_attention}
Vaswani, A., Shazeer, N., Parmar, N., Uszkoreit, J., Jones, L., Gomez, A.N.,
  Kaiser, L.u., Polosukhin, I.: Attention is all you need. In: Guyon, I.,
  Luxburg, U.V., Bengio, S., Wallach, H., Fergus, R., Vishwanathan, S.,
  Garnett, R. (eds.) Advances in Neural Information Processing Systems.
  vol.~30. Curran Associates, Inc. (2017),
  \url{https://proceedings.neurips.cc/paper/2017/file/3f5ee243547dee91fbd053c1c4a845aa-Paper.pdf}

\bibitem{wang2019learning}
Wang, Q., Li, B., Xiao, T., Zhu, J., Li, C., Wong, D.F., Chao, L.S.: Learning
  deep transformer models for machine translation. In: Proceedings of the 57th
  Conference of the Association for Computational Linguistics. pp. 1810--1822.
  Association for Computational Linguistics, Florence, Italy (Jul 2019),
  \url{https://www.aclweb.org/anthology/P19-1176}

\bibitem{wang-etal-2018-switchout}
Wang, X., Pham, H., Dai, Z., Neubig, G.: {S}witch{O}ut: an efficient data
  augmentation algorithm for neural machine translation. In: Proceedings of the
  2018 Conference on Empirical Methods in Natural Language Processing. pp.
  856--861. Association for Computational Linguistics, Brussels, Belgium
  (Oct-Nov 2018). \doi{10.18653/v1/D18-1100},
  \url{https://aclanthology.org/D18-1100}

\bibitem{wei2020multiscale}
Wei, X., Yu, H., Hu, Y., Zhang, Y., Weng, R., Luo, W.: Multiscale collaborative
  deep models for neural machine translation. In: Proceedings of the 58th
  Annual Meeting of the Association for Computational Linguistics. pp.
  414--426. Association for Computational Linguistics, Online (Jul 2020),
  \url{https://www.aclweb.org/anthology/2020.acl-main.40}

\bibitem{wu2019depth}
Wu, L., Wang, Y., Xia, Y., Tian, F., Gao, F., Qin, T., Lai, J., Liu, T.Y.:
  Depth growing for neural machine translation. In: Proceedings of the 57th
  Annual Meeting of the Association for Computational Linguistics. pp.
  5558--5563. Association for Computational Linguistics, Florence, Italy (Jul
  2019). \doi{10.18653/v1/P19-1558},
  \url{https://www.aclweb.org/anthology/P19-1558}

\bibitem{wu-etal-2019-machine}
Wu, L., Zhu, J., He, D., Gao, F., Qin, T., Lai, J., Liu, T.Y.: Machine
  translation with weakly paired documents. In: Proceedings of the 2019
  Conference on Empirical Methods in Natural Language Processing and the 9th
  International Joint Conference on Natural Language Processing (EMNLP-IJCNLP).
  pp. 4375--4384. Association for Computational Linguistics, Hong Kong, China
  (Nov 2019). \doi{10.18653/v1/D19-1446},
  \url{https://aclanthology.org/D19-1446}

\bibitem{xiong2020on}
Xiong, R., Yang, Y., He, D., Zheng, K., Zheng, S., Xing, C., Zhang, H., Lan,
  Y., Wang, L., Liu, T.: On layer normalization in the transformer
  architecture. In: III, H.D., Singh, A. (eds.) Proceedings of the 37th
  International Conference on Machine Learning. Proceedings of Machine Learning
  Research, vol.~119, pp. 10524--10533. PMLR (13--18 Jul 2020),
  \url{https://proceedings.mlr.press/v119/xiong20b.html}

\bibitem{xu-etal-2020-dynamically}
Xu, H., van Genabith, J., Xiong, D., Liu, Q.: Dynamically adjusting transformer
  batch size by monitoring gradient direction change. In: Proceedings of the
  58th Annual Meeting of the Association for Computational Linguistics. pp.
  3519--3524. Association for Computational Linguistics, Online (Jul 2020).
  \doi{10.18653/v1/2020.acl-main.323},
  \url{https://aclanthology.org/2020.acl-main.323}

\bibitem{Neutron}
Xu, H., Liu, Q.: Neutron: An implementation of the transformer translation
  model and its variants. CoRR  \textbf{abs/1903.07402} (2019),
  \url{http://arxiv.org/abs/1903.07402}

\bibitem{xu-etal-2020-lipschitz}
Xu, H., Liu, Q., van Genabith, J., Xiong, D., Zhang, J.: Lipschitz constrained
  parameter initialization for deep transformers. In: Proceedings of the 58th
  Annual Meeting of the Association for Computational Linguistics. pp.
  397--402. Association for Computational Linguistics, Online (Jul 2020).
  \doi{10.18653/v1/2020.acl-main.38},
  \url{https://aclanthology.org/2020.acl-main.38}

\bibitem{xu-etal-2021-optimizing}
Xu, P., Kumar, D., Yang, W., Zi, W., Tang, K., Huang, C., Cheung, J.C.K.,
  Prince, S.J., Cao, Y.: Optimizing deeper transformers on small datasets. In:
  Proceedings of the 59th Annual Meeting of the Association for Computational
  Linguistics and the 11th International Joint Conference on Natural Language
  Processing (Volume 1: Long Papers). pp. 2089--2102. Association for
  Computational Linguistics, Online (Aug 2021).
  \doi{10.18653/v1/2021.acl-long.163},
  \url{https://aclanthology.org/2021.acl-long.163}

\bibitem{zhang2019improving}
Zhang, B., Titov, I., Sennrich, R.: Improving deep transformer with
  depth-scaled initialization and merged attention. In: Proceedings of the 2019
  Conference on Empirical Methods in Natural Language Processing and the 9th
  International Joint Conference on Natural Language Processing (EMNLP-IJCNLP).
  pp. 898--909. Association for Computational Linguistics, Hong Kong, China
  (Nov 2019). \doi{10.18653/v1/D19-1083},
  \url{https://www.aclweb.org/anthology/D19-1083}

\end{thebibliography}

\end{document}